# Leveraging LLMs for Bangla Grammar Error Correction: Error Categorization, Synthetic Data, and Model Evaluation

**Pramit Bhattacharyya** **Arnab Bhattacharya**
Dept. of Computer Science and Engineering,
Indian Institute of Technology Kanpur,
India
pramitb@cse.iitk.ac.in arnabb@cse.iitk.ac.in

## Abstract

Large Language Models (LLMs) perform exceedingly well in Natural Language Understanding (NLU) tasks for many languages including English. However, despite being the fifth most-spoken language globally, Grammatical Error Correction (GEC) in Bangla remains underdeveloped. In this work, we investigate how LLMs can be leveraged for improving Bangla GEC. For that, we first do an extensive categorization of 12 error classes in Bangla, and take a survey of native Bangla speakers to collect real-world errors. We next devise a rule-based noise injection method to create grammatically incorrect sentences corresponding to correct ones. The Vaiyākaraṇa dataset, thus created, consists of 5,67,422 sentences of which 2,27,119 are erroneous. This dataset is then used to instruction-tune LLMs for the task of GEC in Bangla. Evaluations show that instruction-tuning with Vaiyākaraṇa improves GEC performance of LLMs by 3-7 percentage points as compared to the zero-shot setting, and makes them achieve human-like performance in grammatical error identification. Humans, though, remain superior in error correction. The data and code are available from https://github.com/Bangla-iitk/Vaiyakarana.

## 1 Introduction

*Grammatical Error Correction (GEC)* aims to detect and correct grammatical errors in a text automatically. For example, given the following sentence in English, "A ten year oldest boy go to school." a GEC system detects errors in the use of the superlative degree and verb number and corrects it to "A ten-year-old boy goes to school.". The CoNLL 2013 and 2014 shared tasks (Ng et al., 2013, 2014) significantly advanced GEC research, but focus primarily on English.

Bangla (Bengali, বাঙ্গলা/বাংলা Bāṅgalā[1]) is the fifth most spoken language in the world. However, to our knowledge, only a handful of GEC works exist for Bangla. Alam et al. (2007) proposed a rule-based statistical grammar checker, but its coverage of grammatical rules is limited, leading to failure in detecting and correcting complex errors. Data-driven GEC methods require pairs of correct and corresponding incorrect sentences for training. Islam et al. (2018) attempted to generate erroneous Bangla sentences by randomly inserting, deleting, and swapping words in a corpus of 250K sentences. While such methods have been effective for English, they are less suitable for Bangla and other morphologically rich Indian languages with free word order.

Word-level operations such as swapping, deletion, and insertion often fail to produce grammatically incorrect Bangla sentences. Being morphologically rich, Bangla exhibits free word order. Thus, different permutations of subject-verb-object (SVO) are grammatically valid. For instance, consider the Bangla sentence অমর গীতাকে ভালোবাসে। (amara gītākē bhālōvāsē., Amar loves Geeta.) shown in Table A1 of Appx. A. The first five variants contain the same words in different orders, all of which are grammatically correct. The next three variations introduce word substitution, deletion, and insertion; even these forms remain grammatically correct. Thus, a more nuanced error generation approach, beyond simple word-level operations, is needed for Bangla GEC.

Large Language Models (LLMs), particularly those trained with instruction fine-tuning, have demonstrated strong capabilities in generating synthetic data for various NLP tasks, particularly in English (Li et al., 2024; Jin et al., 2024).

In this paper, we explore the ability of LLMs to perform GEC for Bangla. For that, we first do an extensive categorization of possible grammati-cal errors found in Bangla into 12 error categories. We then collected real-world errors by doing an essay writing survey of native speakers. We use the distribution of real errors found from this survey (of 2,576 sentences) to generate synthetic data that contains grammatical errors. The generation is done using a *rule-based noise injection* scheme on grammatically correct sentences to guarantee that the generated sentences are actually grammatically incorrect.

[1] We use ISO15919 transliteration scheme for Bangla: https://en.wikipedia.org/wiki/ISO_15919

We curate a dataset, **Vaiyākaraṇa**[2], consisting of 2,27,119 erroneous and 5,67,422 total Bangla sentences to instruction-tune LLMs. We use the Vaiyākaraṇa dataset to evaluate the ability of LLMs for GEC. We compare the performance of LLMs thus instruction-tuned with Vaiyākaraṇa against both zero-shot performance as well as human evaluators. We also assess their performance for another task, namely, paraphrasing.

Experiments show that using Vaiyākaraṇa improves LLM performance by approximately 3-7 percentage points compared to the zero-shot setting for both GEC and paraphrasing tasks. However, humans still significantly outperform LLMs in correcting grammatically erroneous sentences.

Our key contributions in this paper are:

1. We do an extensive categorization of possible grammatical errors in Bangla into 12 distinct categories (Sec. 3), making this the first such extensive attempt. Standardization of grammatical error categories is essential for addressing challenges in low-resource language (Nigatu et al., 2024). This categorization of errors is applicable for many major Indian languages (Appx. C).
2. We collect and analyze 2,576 human-annotated sentences to identify common grammatical errors. We also present statistics on error distributions based on real-world usage (Sec. 4).
3. We propose a structured error generation approach (Sec. 6) for systematically injecting grammatical errors into Bangla sentences. This approach allows for scalable generation of error-annotated corpora. Our approach is extendable to other Indian languages (Appx. C).
4. We curate Vaiyākaraṇa, a dataset containing 2,27,119 erroneous Bangla sentences, which can be directly used instruction-tuning LLMs for Bangla GEC.
5. We evaluate the effectiveness of instruction-tuned LLMs (decoder-based) for GEC in Bangla against both zero-shot setting and human evaluators (Sec. 7.5). Experiments show that instruction-tuning using Vaiyākaraṇa improves LLM performance by 3-7 percentage points compared to the zero-shot setting for both GEC and paraphrasing tasks. However, humans still significantly outperform LLMs in correcting grammatically erroneous sentences.

[2]The word means "grammarian" in Bangla.

## 2 Related Work

Grammatical Error Correction (GEC) for Indian languages, including Bangla, is still in its early stages. While existing methods have explored rule-based, statistical, and neural approaches, they lack comprehensive error categorization and robust datasets. Early works on GEC focused on rule-based and statistical models. Sonawane et al. (2020) categorized inflectional errors for Hindi GEC, and Rachel et al. (2023) proposed Vyakaranly, a toolkit for Hindi grammar correction. Alam et al. (2007) introduced a rule-based statistical approach, but it failed to generalize beyond simple sentences. Islam et al. (2018) attempted to generate erroneous Bangla sentences via random word swaps, insertions, and deletions. However, this method of generating wrong sentences does not always give the desired result as shown in Table A1 of Appx. A. Rahman et al. (2023) developed a CNN-based spelling correction model, while Oshin et al. (2023) curated a 10K-sentence dataset for Bangla text error classification, with only 3,140 erroneous sentences. Hossain et al. (2023) proposed Panini, a Vaswani-style monolingual transformer for Bangla GEC, and synthetically generated a 7.7M+ sentence corpus over 10 error categories. However, their error classification lacked key categories such as tense errors, Gurucaṇḍālī Dōṣa, and semantic errors, which are significant as reported in Table 2. Hence, we did not use this dataset for instruction-tuning LLMs. Maity et al. (2024) generated a dataset of only 3,412 sentences curated by amalgamating 1,678 sentences (only 50 erroneous samples are publicly available) from essays written by school students and 1,724 sentences by crawling social media websites. This work does not consider number errors, gender errors and semantic errors in Bangla, which may not be significant but do occur occasionally (Sec. 4). Additionally, they do not classify POS and homonym errors, which are significant in Bangla. We have discussed about GEC for English and other languages in Appx B. Back-translation has been used for data augmentation in GEC (Sennrich et al., 2016; Rei and Yannakoudakis, 2017; Zhou et al., 2020). However, round-trip translation for Bangla using English as a bridge fails to consistently produce grammatically incorrect sentences. Moreover, errors generated via back-translation are challenging to categorize and localize, making them unsuitable for Bangla GEC (Sec. 5).

| Error Class | Sub-class | Example of Wrong Sentence (in violet) followed by Correct Sentence (in blue) |
|---|---|---|
| Spelling | Non-Dictionary | আমি কারখানায় কাড করি। (āmi kārakhānāẏa kāḍa kari., I <non-word> in factory.)<br>আমি কারখানায় কাজ করি। (āmi kārakhānāẏa kāja kari., I work in factory.) |
| | Dictionary | আমি কাল বারি যাব। (āmi kāla bāri yāba., I will go water tomorrow.)<br>আমি কাল বাড়ি যাব। (āmi kāla bāṛi yāba., I will go home tomorrow.) |
| | | আমি কাল শাড়ি যাব। (āmi kāla śāṛi yāba., I will go saree tomorrow.)<br>আমি কাল বাড়ি যাব। (āmi kāla bāṛi yāba., I will go home tomorrow.) |
| Word | Tense | আমি গতকাল পড়াশোনা করব। (āmi gatakāla paṛāśōnā karaba., I will study yesterday.)<br>আমি গতকাল পড়াশোনা করেছিলাম। (āmi gatakāla paṛāśōnā karēchilāma., I studied yesterday.) |
| | | যখন শীত আসবে তখন ফুল ফুটেছিল। (yakhana śīta āsabē takhana phula phuṭēchila., When winter comes, flowers bloomed.)<br>যখন শীত আসবে তখন ফুল ফুটবে। (yakhana śīta āsabē takhana phula phuṭabē., When winter comes, flowers will bloom.) |
| | Person | আমি কারখানায় কাজ করে। (āmi kārakhānāẏa kāja karē., I works in factory.)<br>আমি কারখানায় কাজ করি। (āmi kārakhānāẏa kāja kari., I work in factory.) |
| | Number | আমি এখানে চারজন থাকি। (āmi ēkhānē cārajana thāki., I four stay here.)<br>আমরা এখানে চারজন থাকি। (āmarā ēkhānē cārajana thāki., We four stay here.) |
| | Gender | উত্তম একজন অসাধারণ অভিনেত্রী। (uttama ēkajana asādhāraṇa abhinētrī., Uttam is an outstanding actress.)<br>উত্তম একজন অসাধারণ অভিনেতা। (uttama ēkajana asādhāraṇa abhinētā., Uttam is an outstanding actor.) |
| | Case | আমি রান্নাঘরকে ভাত খাই। (āmi rānnāgharakē bhāta khāi., I eat rice to kitchen.)<br>আমি রান্নাঘরে ভাত খাই। (āmi rānnāgharē bhāta khāi., I eat rice in kitchen.) |
| | Parts-of-Speech | হিমালয়ের সুন্দর অবিস্মরণীয়। (himālaẏēra sundara abismaraṇīẏa., The beautiful of Himalaya is unforgettable.)<br>হিমালয়ের সৌন্দর্য অবিস্মরণীয়। (himālaẏēra saundarya abismaraṇīẏa., The beauty of Himalaya is unforgettable.) |
| | Missing | আমি কাল বাড়ি •। (āmi kāla bāṛi •., I • home tomorrow.)<br>আমি কাল বাড়ি যাব। (āmi kāla bāṛi yāba., I will go home tomorrow.) |
| | | উত্তম একজন অসাধারণ •। (uttama ēkajana asādhāraṇa •., Uttam is an outstanding •.)<br>উত্তম একজন অসাধারণ অভিনেতা। (uttama ēkajana asādhāraṇa abhinētā., Uttam is an outstanding actor.) |
| Gurucaṇḍālī dōṣa | | নন্দবাবু ইহা লক্ষ্য করেছেন। (nandabābu ihā lakṣya karēchēna., Nanda has noticed this.)<br>নন্দবাবু ইহা লক্ষ্য করিয়াছেন। (nandabābu ihā lakṣya kariẏāchēna., Nanda has noticed this.) |
| | | নন্দবাবু ইহা লক্ষ্য করেছেন। (nandabābu ihā lakṣya karēchēna., Nanda has noticed this.)<br>নন্দবাবু এটা লক্ষ্য করেছেন। (nandabābu ēṭā lakṣya karēchēna., Nanda has noticed this.) |
| Punctuation | | আমি গতকাল পড়াশোনা করেছিলাম? (āmi gatakāla paṛāśōnā karēchilāma?, I studied yesterday?)<br>আমি গতকাল পড়াশোনা করেছিলাম। (āmi gatakāla paṛāśōnā karēchilāma., I studied yesterday.) |
| Semantic | | মানস আকাশ খেতে ভালোবাসে। (mānasa ākāśa khētē bhālōbāsē., Manas loves to eat the sky.)<br>মানস আকাশ দেখতে ভালোবাসে। (mānasa ākāśa dēkhatē bhālōbāsē., Manas loves to see the sky.) |
| | | মানস আকাশ খেতে ভালোবাসে। (mānasa ākāśa khētē bhālōbāsē., Manas loves to eat the sky.)<br>মানস মাছ খেতে ভালোবাসে। (mānasa mācha khētē bhālōbāsē., Manas loves to eat fish.) |

Table 1: Grammatical Error Types in Bangla

Since Bangla lacks large-scale GEC datasets, we adopt a structured noise injection approach based on real-world error patterns, ensuring the controlled generation of incorrect sentences across specific error types. This approach improves sentence quality while maintaining category-wise error distributions observed in real time. Unlike previous methods, our dataset explicitly covers common Bangla errors and is adaptable to other Indian languages (Appx. C).

## 3 Grammar Error Categories

Standardization of error categories is necessary for GEC to alleviate problems associated with low-resource languages (Nigatu et al., 2024). In this section, we categorize grammatical errors in Bangla formally. We follow a standard Bangla grammar book (Chakroborty, 2018) as reference for grammar error types. The book explicitly does not have any error categories; however, following the grammatical rules described in the book, we have formalised these error categories. To our humble understanding, these categories are exhaustive and cover all possible error types in Bangla. They can be also used for other major Indian languages (Appx. C).

First, we classified the grammatical errors in Bangla into 5 broader categories. These broad categories are then further sub-divided into 12 finer distinctions. Table 1 lists example sentences for all the categories[3] of the error classes. Appx. D explains the category of errors in detail with examples. A sentence may contain multiple errors of one class or different classes.

[3]The text in violet shows the erroneous portion of a sentence corresponding to the correct text in blue.

## 4 Manual Generation

To understand the nature of real-life grammatical errors made by native speakers, we organized a survey in which participants were asked to write an essay on a specific topic within a specific time. In this way, we collected handwritten sentences from native Bangla speakers and analyzed the occurrence of various grammatical errors in those sentences. Each participant was allowed 30 minutes to write an essay comprising at least 15 sentences and 150 words, choosing from a set of topics provided. The survey was conducted in a proctored environment to simulate an exam-like situation, allowing us to collect real-time data (including errors) on Bangla writing. We collected 123 essays, resulting in 2,576 sentences and 28,713 words, produced by 51 participants (30 participants wrote 2 essays each, while 21 participants wrote 3 essays). The longest sentence contained 69 words, while the shortest had just 1 word. Detailed information about the topics and participants can be found in Appx. E. A team of 3 Bangla language experts then evaluated the written sentences and the errors were categorised based on majority voting. Of the 2,576 sentences written, 1,045 (41%) were grammatically incorrect.[4] Of the erroneous sentences, 804 (77%) contained single errors, while 241 (23%) had multiple errors. Among the sentences with multiple errors, 185 contained 2 errors (18%), and 42 sentences contained 3 errors (4%). The remaining 9 sentences contained up to 6 errors. A total of 678 (4%) words were erroneous in these 1,045 sentences.

Table 2 presents the number of errors for each category outlined in Table 1, along with their respective percentages for the total number of erroneous words. Spelling mistakes emerged as the most prevalent type of error, representing over 62% of occurrences. Further analysis indicates that more than 45% of spelling errors stem from the confusion between the characters 'ন'(n)/'ণ'(ṇ); 'র'(r)/'ড়'(ṛ)/'ঢ়'(ṛh); and 'স'(s)/'শ'(ś)/'ষ'(ṣ). The most significant dictionary-based spelling error involves the mixing of "কি" (ki, whether) and "কী" (kī, what). Although tense, person, number, gender, and semantic errors are not as frequent in Bangla, they do occur once in a while. Case and punctuation errors are also rather common. Additionally, we observed that sentences containing multiple errors generally include several spelling mistakes. Notably, the combination of (spelling, punctuation) and (spelling, case, and punctuation) errors occurs the most. In one case, as many as six different types of errors (spelling errors of both kinds, case, missing word, Gurucaṇḍālī Dōṣa, and punctuation) appeared in a single sentence. We next use the relative frequency of these different kinds of errors for our synthetic data generation.

| Error Class | | #Occurrences | Percentage |
|---|---|---|---|
| | Non-Dictionary | 677 | 49.85 |
| | Dictionary | 174 | 12.81% |
| Spelling Errors | | 851 | 62.66% |
| | Tense Errors | 30 | 2.21% |
| | Person Errors | 26 | 1.91% |
| | Number Errors | 4 | 0.29% |
| | Gender Errors | 1 | 0.07% |
| | Case Errors | 162 | 11.93% |
| | POS Errors | 29 | 2.14% |
| | Missing Words | 64 | 4.71% |
| Word Errors | | 316 | 23.26% |
| Punctuation Errors | | 156 | 11.49% |
| Semantic Errors | | 2 | 0.15% |
| Gurucaṇḍālī Dōṣa | | 33 | 2.43% |
| **Total** | | 1,358 | 100.00% |

Table 2: Grammatical errors in manual survey

## 5 Error Injection Methods

In this section, we focus on generating synthetic data mimicking the real-world error distribution. Bryant et al. (2023) proposed various synthetic data generation methods, including back-translation (Sennrich et al., 2016), round-trip translation (Zhou et al., 2020) and error injection methods (Bryant et al., 2023) using grammatical methods. We next show the advantages and disadvantages of each type of methodology for synthetic data generation for Bangla GEC.

### 5.1 Translation Methods

We start by evaluating back translation and round-trip translation methods to see if they can generate erroneous sentences. Back translation involves translating another language (here, English) text to Bangla. This method introduces subtle variations that mimic realistic errors in text, creating a diverse dataset for model training. However, this method does *not* always guarantee the genera-tion of erroneous sentences, as shown in Table A3 of Appx. F. Zhou et al. (2020) used round-trip translation, a variant of back-translation, to synthesize noisy sentences using a bridge language, e.g., English-Chinese-English, where Chinese is the bridge language. We tried to generate the wrong sentences by following the same methodology using English as a bridge language. Example sentences are shown in Table A4 of Appx. F.

[4]This is likely due to the exam-like time situation where participants did not get a time to revise and correct.

### 5.2 LLM-generated Sentences

The same phenomenon is observed while trying to generate erroneous sentences using GPT-4o. We used the prompt ব্যাকরণগত ভুল বাক্য লেখ (byākaraṇagata bhula bākya lēkha, Write a grammatically wrong sentence). While it generates sentences such as আমি কাল রাতি সিনেমা দেখছিলাম। (āmi kāla rāti sinēmā dēkhachilāma., I saw a cinema last <non-word>.) which are grammatically wrong, it also generates sentences like তারা ফুটবল খেলেছে গতকাল। (tārā phuṭabala khēlēchē gatakāla., They played football yesterday.) which are grammatically correct. Other methods, including adding paraphrased sentences also suffers from similar issues.

From the above discussion, it is clear that none of these methods guarantees a generation of grammatically wrong sentences. Further, it does not categorize and localize errors in the generated sentences. Hence, we adopted the method of error injection following grammatical rules as the preferred method of generating erroneous sentences.

### 5.3 Rule-based Error Injection

Through our rule-based error injection methodology, not only can we guarantee the presence of grammatical errors in a sentence but also categorize and localize the type of error in the sentence. We introduced noise into grammatically correct sentences following the error distribution shown in Table 2, and curated Vaiyākaraṇa with 2,27,119 erroneous sentences. The total number of sentences is 5,67,422.

For errors related to homonyms, parts-of-speech (POS), tense, person, and case, we primarily replaced the correct word with a corresponding incorrect word (all such pairs were sourced from (Chakroborty, 2018)). In this substitution process, we ensured that for generating tense at least two verbs are present in the sentence. Similarly, for Gurucaṇḍālī Dōṣa errors, at least two verb/pronoun are present. We intentionally do not replace one of these words to ensure the generation of an erroneous sentences. Since we inject noise according to the rules outlined in the book (Chakroborty, 2018), the resulting sentences are *guaranteed* to be *always incorrect*. The detailed steps taken to introduce noise for various types of errors are described in Appx. G.

## 6 Instruction-tuning Dataset: Vaiyākaraṇa

To leverage LLMs for Bangla GEC, we curate a corpus of grammatically incorrect sentences by following the rule injection methodology as described in the previous section.

We use the Vācaspati (Bhattacharyya et al., 2023) corpus as the base set of grammatically correct sentences. We chose this corpus since it consists of only literature data and, hence, sentences sampled from this corpus are grammatically correct, as reported by the authors. Further, the corpus captures stylistic, linguistic, spatial and temporal variations of Bangla. The temporal diversity, in particular, is especially useful for Gurucaṇḍālī Dōṣa errors since that particular writing style started becoming rare from 1960s. (Newspapers, blogs, and social media data are not suitable for this.)

We collected 5,67,172 sentences from Vācaspati, which serve as our gold-standard sentences. Additionally, we incorporated 250 sentences from a well-known grammar book (Chakroborty, 2018) to enhance the grammatical and linguistic richness of the dataset. These sentences are beneficial for generating errors related to number, gender, and semantics according to the rules specified in the book, which are less frequently found in literary data. Consequently, we curated a dataset containing a total of 5,67,422 sentences. We followed the data cleaning and pre-processing steps outlined in Appx. H to ensure that this dataset is suitable for error generation.

We inject errors into randomly selected sentences following grammatical rules, with the type of error and the words to inject errors chosen randomly. We choose roughly 40% sentences for error injection. This process ensures a diverse range of errors. Since most real-world erroneous sentences typically involve single-word mistakes, we generate sentences that reflect this pattern. To cover various error categories, we carefully monitor the number of erroneous words to

| Error Class | #Occurrences | Total % | Error % |
|---|---|---|---|
| Non-Dictionary | 113,244 | 19.97 | 49.91 |
| Dictionary | 26,046 | 4.59 | 11.48 |
| Spelling Errors | 139,290 | 24.56 | 61.40 |
| Tense Errors | 4,983 | 0.88 | 2.20 |
| Person Errors | 4,530 | 0.80 | 2.00 |
| Number Errors | 200 | 0.035 | 0.090 |
| Gender Errors | 100 | 0.018 | 0.044 |
| Case Errors | 26,046 | 4.59 | 11.48 |
| POS Errors | 4622 | 0.81 | 2.04 |
| Missing Words | 10,690 | 1.88 | 4.71 |
| Word Errors | 51,171 | 9.02 | 22.55 |
| Punctuation Errors | 26,046 | 4.59 | 11.48 |
| Semantic Errors | 100 | 0.018 | 0.044 |
| Gurucaṇḍālī Dōṣa | 6,402 | 1.13 | 2.82 |
| Multiple Errors | 3,860 | 0.68 | 1.70 |
| *InCorrect* | 2,26,869 | 40.00 | 100.00 |
| *Correct* | 340,303 | 60.00 | 0.00 |
| **Total** | 5,67,422 | 100.00 | 100.00 |

Table 3: Grammatical Error in Vaiyākaraṇa ensure it does not exceed 30% of the total words in any given sentence. Since most real-world spelling errors arise from confusion between characters such as 'ন'(n)/'ণ'(ṇ); 'র'(r)/'ড়'(ṛ)/'ঢ়'(ṛh); 'স'(s)/'শ'(ś)/'ষ'(ṣ); "কী"(kī)/"কি"(ki) and other similar sounding words, we place additional emphasis on sentences containing these characters or words when introducing spelling errors. For all other types of mistakes, errors are injected randomly with equal probabilities. The detailed steps taken to introduce noise for various types of errors are described in Appx. G.

Table 3 shows the number of sentences generated and their corresponding category of errors. A total of 2,23,259 sentences contains a single error, whereas 3,860 sentences contain multiple errors. Table A6 in Appx. J shows the distribution of multiple errors in Vaiyākaraṇa. The maximum number of errors in a sentence in Vaiyākaraṇa is 10. We validated that the distribution of error categories are comparable between real-world data and generated data (Appx I).

Sec. 4 showed that ∼77% of sentences contain single errors, while the remaining contain multiple errors. Amongst them, the most common is multiple spelling errors. If there are multiple errors of the same category in a sentence, we have categorized it in that error category only, and not included them into the multiple error category. Table A6 shows the distribution of sentences with multiple error types in Vaiyākaraṇa. We have considered only those category of multiple errors that were present in the human survey in Sec. 4, and not all possible combinations.

| Correct | Generated | Error Type |
|---|---|---|
| কেউ উঁকিঝুঁকি মারে না, ডিস্টার্ব করতে নামে না। (kēu um̐kijhum̐ki mārē nā, ḍisṭārva karatē nāmē nā., No one peeks, no one comes to disturb.) | কেউ উঁকিঝুঁকি মারে না, ডিস্টার্ব করিতে নামে না। (kēu um̐kijhum̐ki mārē nā, ḍisṭārva karitē nāmē nā.) | Gurucaṇḍālī Dōṣa |
| পল্টুর যে ফাঁসির হুকুম হয়েছে এটা সে নবীনকে বলেনি। (palṭura yē phām̐sira hukuma haẏēchē ēṭā sē nabīnakē balēni., He did not tell Navin that Paltu has been sentenced to death.) | পল্টুর যে ফাঁশির হুকুম হয়েছে এটা সে নবীনকে বলেনি। (palṭura yē phām̐śira hukuma haẏēchē ēṭā sē nabīnakē balēni.) | Non-Dictionary |
| অন্ধকারে গাছপালা ভেদ করে সে কি ছুট! (andhakārē gāchapālā bhēda karē sē ki chuṭa!, What a run through the trees in the dark!) | অন্ধকারে গাছপালা ভেদ করে সে কী ছুট! (andhakārē gāchapālā bhēda karē sē kī chuṭa!) | Dictionary |

Table 4: Example of sentences generated by noise-injection method present in Vaiyākaraṇa.

Although we focused on generating a GEC dataset Vaiyākaraṇa for Bangla, the aforementioned procedure of injecting noise to generate grammatically wrong sentences can also be applied to other major Indian languages with little or no modification (Appx. C).

We next show some anecdotal examples in Table 4 to highlight that our error injection method generates grammatically incorrect sentences that resemble real-life ones. It covers homonym errors as well as the popular confusion of শ (ś) / স (s).

Additionally, we validated these sentences by a group of 12 individuals. Each participant was provided with a unique set of 50 sentences and was asked to mark whether they believed the errors were naturally occurring, as well as the minimum education level required to detect the errors. Of the 12 participants, 10 felt that the generated sentences were natural and that a minimum education level of 10th grade would be sufficient to identify the errors. The remaining 2 participants noted that the generated sentences sometimes seemed artificial. Participants also rated the level of naturalness on a Likert scale from 1 (least natural) to 5 (most natural), and the average score for the generated sentences was 3.62.

## 7 Evaluation

In this section, we evaluate the effectiveness of our rule-based noise-injection method and the instruction-tuned dataset Vaiyākaraṇa by evaluating large language models (LLMs) on grammatical error detection and correction, as well as on a separate task of paraphrasing.

### 7.1 Efficacy of the Rule-based Method

We prompted GPT-4o with the instruction "ব্যাকরণগত ভুল বাক্য লেখ" (byākaraṇagata bhula bākya lēkha, Write a grammatically incorrect sentence) and analyzed the generated outputs. From a set of 50 generated sentences, each containing less than five words, we observed that it generated 9 *correct* sentences and 41 *incorrect* sentences. Among the 41 incorrect sentences, 32 sentences were of Tense and Person errors, whereas the rest are Spelling errors. Notably, although spelling errors are generally the most frequent error type in Bangla, GPT-4o generated a disproportionate number of tense and person errors, ignoring all the other categories of grammatical errors.

We repeated the same experiment with a minimum word length constraint of at least 8. We observed that GPT-4o generated 11 *correct* sentences and 39 *incorrect* sentences. All 39 incorrect sentences contained either Tense or Person errors only. Additionally, GPT-4o misclassified 3 incorrect sentences as grammatically correct.

We further experimented with back-translation and round-trip translation for generating erroneous sentences. While 23 out of 50 sentences in the back translation were grammatically correct, the corresponding number for round-trip translation was even higher—40. The common error types in these cases were Spelling (non-dictionary words), Tense, and Person while other error categories were practically non-existent.

These experiments indicate that LLMs like GPT-4o struggle to generate grammatically incorrect sentences reliably. Even when incorrect sentences are generated, the types of errors are limited in diversity, and are predominantly tense and person errors. In contrast, our proposed rule-based noise injection approach involves injecting errors based on a set of predefined rules, ensuring a more diverse range of error types. Also, it guarantees generation of incorrect sentences. Further, even when LLM-generated sentences are erroneous, categorizing and localizing the errors remains a significant challenge. This validates the necessity and effectiveness of our controlled rule-based error generation framework for Bangla.

### 7.2 Grammatical Error Detection and Classification

We first test the abilities of humans and LLMs on error classification. We have segregated this into three types. The first is *binary* classification, where the task is to indicate if a given sentence is grammatically correct or wrong. The next two tasks are on classifying an erroneous sentence into the type of error it has. Assuming the correct sentence to be another class, the task is to either classify broadly into 5+1 broad classes or in a fine manner into 12+1 classes.

For human evaluation, we developed an interface where participants could mark the error class for a given sentence, including the option for the correct classification. We enlisted the help of 12 Bangla speakers, each of whom was assigned a set of 50 non-overlapping sentences. These sentences were randomly selected from a pool of 2,500 sentences from the Vaiyākaraṇa dataset, comprising 650 correct sentences and 1,850 incorrect ones. A very high score on the part of humans would indicate that the dataset is very easy to classify, and may not be realistic. Otherwise, the wrong sentences generated by our noise injection method would be realistic and non-trivial.

Table 5 shows the results of encoder-decoder based transformer models against humans for the three kinds of classification tasks on these 600 sentences. The mean macro-F1 scores achieved by the 12 evaluators for the three classification tasks were between 82% and 89%. The highest macro-F1 scores recorded by an individual evaluator were 91.10%, 87.50%, and 83.33% for the binary, broad and fine classification tasks respectively. The high macro-F1 scores indicate that humans can identify and categorize the errors fairly well, which indicates that the synthetic data is following the naturally occurring error trends. However, the scores are not very high, thereby implying the test sentences are realistic and not trivial to correct.

These results do not include decoder-based models since they are known to be not good for classification tasks (Nielsen et al., 2025). Nevertheless, we tried using them for the simplest task—that of binary classification. Results shown in Table A8 in Appx. K confirm that their performance is indeed not up to the mark. Hence, we have not continued with our evaluation for multi-class scenarios using decoder-based models.

| Model | Parameters | Binary | Broad | Fine |
|---|---|---|---|---|
| Google-ByT5 | 300M | 81.25±0.65 | 79.30±0.87 | 76.65±0.80 |
| BanglaT5 | 277M | 88.90±0.10 | 84.50±0.68 | 82.48±0.14 |
| Panini | 70.46M | 89.25±0.25 | 84.75±0.15 | 82.88±0.08 |
| Human | – | 88.30±3.46 | 84.20±3.75 | 82.30±3.90 |

Table 5: Macro-F1 for error classification on 600 sentences.

For each of the other transformer models, we employed 5-fold cross validation. When the transformers are tested on a zero-shot setting, they failed to perform any classification, and simply classified every sentence into 1 class. Fine-tuning these same models with our Vaiyākaraṇa dataset makes them achieve results that are at par with humans. Panini achieves the highest mean score and the best macro-F1 score for both multi-class and binary classification tasks. On average, both Panini and Bangla-T5 outperform human evaluators, although the maximum scores achieved by individual humans are higher.

We next evaluated the performance of various neural models including multilingual LLMs such as GPT-4o (OpenAI et al., 2023), GPT-2-XL (Black et al., 2022), Bloom-1.1B (Workshop et al., 2023), BanglaT5 (Hossain et al., 2023), and Google-ByT5. Each model was run five times for at least 20 iterations, utilizing different seed values, and we report the means and standard deviations in Table 6. Details regarding the hyperparameters for all models can be found in Appx. M.

### 7.3 Grammatical Error Correction

We now investigate the ability of humans and neural models to correct grammatically wrong sentences. In this section, we have only considered neural models that can generate sentences; hence, encoder-only models such as BanglaBERT, VĀCBERT, etc. have been ignored.

We have conducted another human survey with 12 people (5 persons are overlapping with the previous set of 12 annotators) and 600 sentences (the same sentences used for LLM evaluation). Each annotator was given a set of 50 random sentences and was asked to mark them as right or wrong. In addition, if they felt a sentence was wrong, they were asked to provide a correct variant of the sentence with as few changes to the original sentence as possible. The maximum score achieved by a human was 47 out of 50 (94%) with an average of 39.3 (78.6%) and a standard deviation of 5.66. Most corrected variants (∼92%) matched the gold standard sentence. However, some correct answers deviated more, and the maximum deviation was 4 words (the average sentence length is around 10 words).

| Model | Zero-Shot | | | | With Fine-tuning | | | |
|---|---|---|---|---|---|---|---|---|
| | GLEU | $F_{0.5}$ | BERT-score | BLEU | GLEU | $F_{0.5}$ | BERT-score | BLEU |
| GPT-4o | 70.30 | 60.25 | 81.85 | 46.40 | 73.66 | 63.57 | 86.86 | 51.45 |
| GPT-2-XL | 62.85 | 58.03 | 77.86 | 42.90 | 69.35 | 60.27 | 83.60 | 50.10 |
| BLOOM-1.1B | 61.45 | 54.43 | 75.86 | 40.86 | 66.50 | 57.77 | 79.24 | 46.60 |
| BanglaT5 | 62.55 | 54.00 | 76.82 | 41.67 | 68.80 | 58.70 | 81.50 | 48.90 |
| Google-ByT5 | 61.20 | 54.00 | 74.45 | 40.50 | 66.10 | 57.20 | 80.20 | 46.45 |
| Panini | 64.90 | 56.17 | 76.90 | 40.80 | 71.10 | 60.50 | 81.70 | 50.35 |
| LLaMA3-8B | 66.52 | 59.13 | 79.84 | 44.63 | 71.49 | 61.91 | 85.22 | 50.77 |

Table 6: Performance of models with and without fine-tuning with Vaiyākaraṇa on grammatically correct sentence generation on 52,100 generated sentences.

We created a test corpus of 52,100 sentences (details of the sentences are in Appx. L) which were *not* present in Vaiyākaraṇa and tested the ability of the models to generate correct grammatical sentences in two conditions, one without instruction tuning (zero-shot), and the other after instruction tuning the models with Vaiyākaraṇa. The hyperparameters used for instruction-tuning the models are described in Appx. N. Table 6 shows the performance of models in generating grammatically correct sentences with and without Vaiyākaraṇa. There is an average increase of more than 5% on $F_{0.5}$ score as well as GLEU. In all the architectures, the average $F_{0.5}$ and GLEU increased from the zero-shot paradigm. It indicates the effectiveness of Vaiyākaraṇa in building better GEC models for Bangla. In addition to the standard GEC metric GLEU, we used the other metrics since they are proposed in the CoNLL 2013 task (Ng et al., 2013) and Panini (Hossain et al., 2023).

To evaluate the quality of Vaiyākaraṇa, we experimented with generating grammatically correct sentences with generative models. In this experiment, we tested these models' performance on 2,576 manually written sentences obtained from essay writing surveys that were not included in Vaiyākaraṇa. Like our previous experiments, we assessed the generative models in zero-shot and after instruction-tuning with Vaiyākaraṇa. Table 7 demonstrates that the performance of all models and architectures improved by 3-5 percentage points after instruction-tuning with Vaiyākaraṇa. Thus, the same trends hold across both the generated and the manual sentences.

Fig. 1 shows the GLEU score for each of the 12 error categories for GPT-4o on the 1513 manually written sentences. Fig. 3 of Appx. L, on the other hand, shows the GLEU score for each of the

| Model | Zero-Shot | | | | With Fine-tuning | | | |
|---|---|---|---|---|---|---|---|---|
| | GLEU | $F_{0.5}$ | BERT-score | BLEU | GLEU | $F_{0.5}$ | BERT-score | BLEU |
| GPT-4o | 72.30 | 60.25 | 82.35 | 48.50 | 75.66 | 64.60 | 89.90 | 55.15 |
| GPT-2-XL | 64.85 | 58.53 | 70.16 | 44.50 | 70.85 | 61.30 | 84.65 | 51.50 |
| BLOOM-1.1B | 63.75 | 53.85 | 77.56 | 43.30 | 68.60 | 58.40 | 81.35 | 48.80 |
| BanglaT5 | 65.75 | 58.20 | 79.35 | 45.70 | 68.50 | 59.80 | 83.60 | 51.40 |
| Google-ByT5 | 62.60 | 55.00 | 76.15 | 42.75 | 66.90 | 58.50 | 81.10 | 48.25 |
| Panini | 66.80 | 57.40 | 79.50 | 47.40 | 71.20 | 72.30 | 85.00 | 52.25 |
| LLaMA3-8B | 68.52 | 59.39 | 76.13 | 46.48 | 73.24 | 62.94 | 87.26 | 53.31 |

Table 7: Performance of models with and without fine-tuning with Vaiyākaraṇa on grammatically correct sentence generation on 2,576 manual sentences.

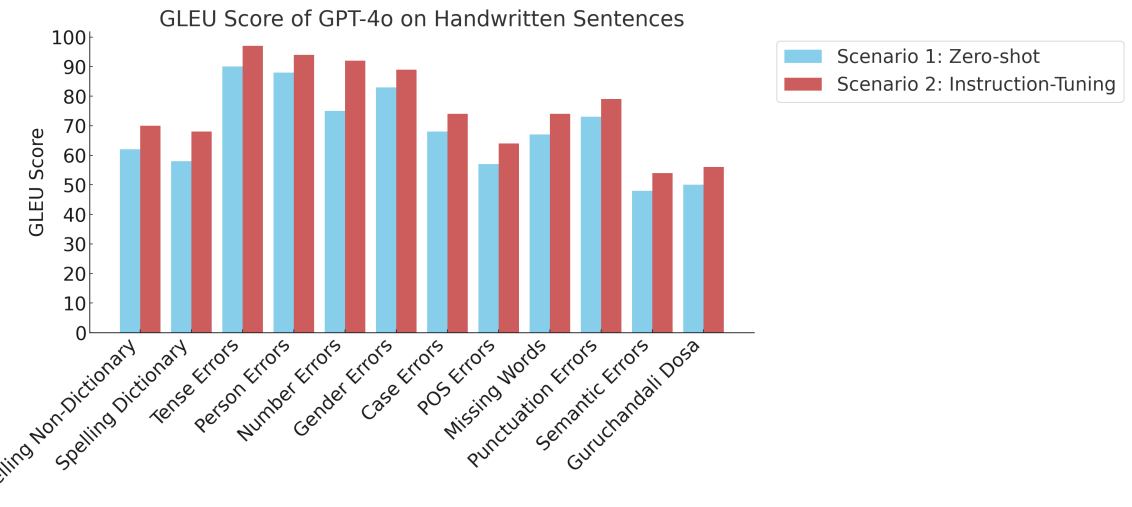

Figure 1: Performance of GPT-4o on different error categories in Bangla on 2,576 manual sentences.

12 error categories for the best performing model, GPT-4o, on the 20,100 sentences used for GEC evaluation of transformer models.

Appx O shows qualitative examples of performance of human and GPT-4o for GEC in Bangla.

### 7.4 Performance of LLMs on Error Categories

We evaluated the performance of GPT-4o (which is the best-performing model) for each error class after instruction tuning with different sizes for each category on the Vaiyākaraṇa dataset. The detailed performance of GPT-4o for each error class is shown in Fig. 2. It indicates that gpt4-o struggles to correct Spelling errors (specially Homonym errors), Case errors, POS errors, and Gurucaṇḍālī Dōṣa. It fairly struggles in correcting Missing Word and Punctuation and Semantic errors, while it is quite good in correcting Number, Gender, Tense and Person errors.

### 7.5 Paraphrasing

To demonstrate the quality of Vaiyākaraṇa as a corpus, we evaluated the performance of neural models on a completely different task, that of paraphrasing, before and after instruction-tuning with Vaiyākaraṇa. We used the paraphrasing dataset developed by (Akil et al., 2022) containing 5,763 sentences. Table 8 shows the performance of models with and without instruction-tuning using Vaiyākaraṇa. The models' performances were enhanced by 3-7 percentage points after instruction-tuning with Vaiyākaraṇa.

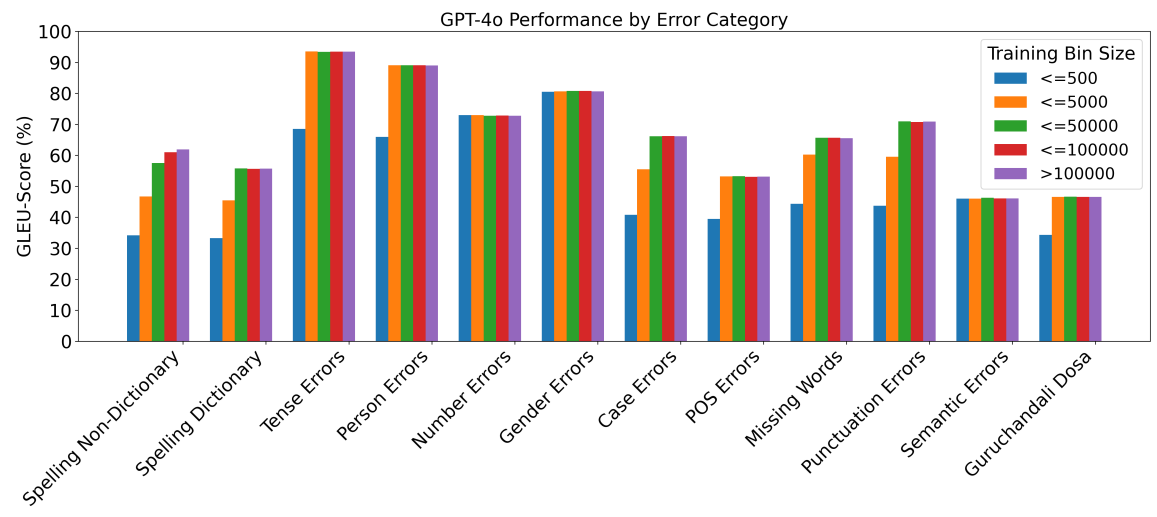

Figure 2: Performance of GPT-4o on different error categories on different input sentences for that category.

| Model | Zero-Shot | | With Fine-tuning | |
|---|---|---|---|---|
| | BLEU | BERT-score | BLEU | BERT-score |
| GPT-4o | 49.00 | 62.70 | 57.60 | 66.80 |
| GPT-2 | 46.38 | 57.58 | 50.74 | 62.41 |
| BLOOM | 45.50 | 55.86 | 49.00 | 59.50 |
| BanglaT5 | 48.66 | 57.97 | 55.38 | 62.53 |
| Google-ByT5 | 45.88 | 56.52 | 56.35 | 61.42 |
| Panini | 48.46 | 57.50 | 55.56 | 62.75 |
| LLaMA3-8B | 47.68 | 60.11 | 54.12 | 64.59 |

Table 8: Performance of models with and without fine-tuning with Vaiyākaraṇa on paraphrasing

## 8 Conclusions and Future Work

In this paper, we proposed a rule-based noise injection methodology for generating grammatically wrong sentences in Bangla. We generated erroneous sentences across 12 categories, which is the most extensive categorization of grammatical errors for Bangla. We curated a dataset Vaiyākaraṇa consisting of 2,27,119 wrong and 5,67,422 total sentences. We also collected a set of 2,576 sentences (of which 1,045 were grammatically wrong) from manually written essays. The results show that neural models perform similar to human evaluators in detecting error categories and words whereas humans outperform these models in correcting grammatically wrong sentences. After instruction-tuning with Vaiyākaraṇa, the performance of LLMs improve by 3-7 percentage points on both these tasks. We have released Vaiyākaraṇa, the Alpaca format of Vaiyākaraṇa, manual hand-written data and code for the rule-based noise injection methodology of the paper under a non-commercial license at https://github.com/Bangla-iitk/Vaiyakarana.

In future, we would like to use this dataset to develop better GEC models with explainability. Further, this methodology can be applied to generate benchmarks for most other major Indian languages, since their grammatical structures are similar to Bangla.

## 9 Limitations

Curating a large quality benchmark for GEC requires a good quality lemmatizer and POS tagger. Bangla suffers from a lack of quality lemmatizers and POS taggers. Hence, we had to manually add words from a grammar book (Chakroborty, 2018).

Also, hand-written Bangla data is not readily available. We conducted a survey for several weeks and could still collect only 2,576 hand-written sentences.

Finally, while the 12 human evaluators are all native speakers of Bangla, evaluating against Bangla grammarians could have given us more insights into the process. We are planning to do that in the future.

Additionally, we have evaluated only GPT-4o and not other commercially available models like Claude and Gemini due to resource constraints. In contrast, we have experimented with different architectures and loss functions to show the generalizability of the results.

## 10 Ethics Statement

The Vaiyākaraṇa benchmark is curated by merging sentences from Vacaspati corpus (Bhattacharyya et al., 2023) and (Chakroborty, 2018). The authors of Vacaspati provided us with the corpus, and (Chakroborty, 2018) is publicly available. Hence, there is no copyright infringement in curating Vaiyākaraṇa. We have made efforts to ensure that Vaiyākaraṇa is also devoid of any objectionable statements. We have also conducted a manual essay writing survey for gathering real word errors. The participants have kindly allowed us to use their essays for research purpose.

## Acknowledgements

We thank *Arghya Mukherjee*, a PhD student in the department of Mathematics and Statistics, IIT Kanpur, for his unconditional support that played a pivotal role in the completion of the project. We also acknowledge the *SERB Core Research Grant* (file no. CRG/2023/006064) and the *Research-I Foundation (RIF)*, Dept. of Computer Science and Engineering, IIT Kanpur grants, which were essential for completion of the project.

## Appendix

## A Word Order

Table A1 shows that all possible word order of sentence অমর গীতাকে ভালোবাসে। (amara gītākē bhālōvāsē.) is correct.

| | | |
|---|---|---|
| Original sentence | অমর গীতাকে ভালোবাসে। amara gītākē bhālōvāsē. | |
| Word order 1 | গীতাকে অমর ভালোবাসে। gītākē amara bhālōvāsē. | |
| Word order 2 | গীতাকে ভালোবাসে অমর। gītākē amara bhālōvāsē. | |
| Word order 3 | অমর ভালোবাসে গীতাকে। gītākē amara bhālōvāsē. | |
| Word order 4 | ভালোবাসে অমর গীতাকে। gītākē amara bhālōvāsē. | |
| Word order 5 | ভালোবাসে গীতাকে অমর। gītākē amara bhālōvāsē. | |
| Word substitution | শ্যামল গীতাকে ভালোবাসে। śyāmala gītākē bhālōvāsē. | |
| Word deletion | অমর ভালোবাসে। amara bhālōvāsē. | |
| Word insertion | অমর গীতাকে খুব ভালোবাসে। amara gītākē khuva bhālōvāsē. | |

Table A1: Word order shuffling, substitution, deletion and insertion may not necessarily generate wrong sentences in Bangla.

## B Related Work

In this section, we discuss GEC in English and other non-Indian languages.

**English**: CoNLL-shared task 2013(Ng et al., 2013) and CoNLL-shared task 2014(Ng et al., 2014) played a pivotal role in advancing GEC works in English. Other than providing 55,000+ grammatically incorrect sentences in English, they also categorized grammatical errors in English into 5 broad classes and 27 finer classes. Napoles et al. (2017) presented a parallel corpus of 1,511 sentences for English representing a wide range of language proficiency. It incorporates holistic edits that make the original text sound more native. Additionally, Yannakoudakis et al. (2011) curated a collection of 1,238 scripts from distinct learners. The BEA-2019 shared task (Bryant et al., 2019) contributed a benchmark dataset of 43,169 sentences curated from the Write&Improve+LOCNESS corpus. This represents a broader range of native English learners.

**Other Languages:** Unlike English, low-resource Asian languages suffer from the unavailability of large corpora for neural models. Attempts have been made to enrich resources for GEC in many languages: Spanish (Davidson et al., 2020), German (Boyd, 2018), Russian (Rozovskaya and Roth, 2019), Czech (Náplava and Straka, 2019), Greek (Korre and Pavlopoulos, 2022), and Chinese (Rao et al., 2018). Syvokon et al. (2023) presented a corpus annotated for GEC and fluency edits for Ukrainian. Lee et al. (2021) gave four different noising methods, such as grapheme-to-phoneme noising rules and, heuristic-based noising rules, and others, to generate incorrect sentences for Korean. Lichtarge et al. (2019) proposed a rule-based system for deliberately injecting noises for low-resource languages like Indonesian (Irmawati et al., 2017). Solyman et al. (2022) proposed semi-supervised noising methods to generate 13,333,929 synthetic parallel examples from a monolingual corpus for Arabic.

## C Human Evaluation for Hindi

We conducted a survey for Hindi grammatical correction to evaluate whether the error injection methodology can be extended to other Indian languages. Five native speakers participated in this survey. Each participant was provided with 40 randomly selected sentences from a pool of 250 sentences (both correct and incorrect). The other setting for this experiment is similar to the Bangla evaluation described in Sec. 7.2.

The average macro-F1 score for all three classification tasks is 83%, 74% and 63%, with the highest being 87%, 87%, and 73.33%. All the participants opined that the generated sentences were confusing and that the sentences identified as incorrect were indeed wrong. This supports our assertion that our method for generating incorrect sentences for Bangla can also be effectively used for other Indian languages, such as Hindi.

## D Grammar Error Categories

This section expands upon Sec. 3 by illustrating the various error categories with examples. Grammatical errors in Bangla can be classified into five broader categories, which are further detailed into twelve specific distinctions. A sentence may contain multiple errors from the same category or from different categories. Table 1 provides example sentences for each of the error classes.

### D.1 Spelling Errors

Spelling errors are amongst the most frequent types of errors. In Bangla and major Indian languages, there are almost similar sounding consonants and, thus, mistakes between ন / ণ (n / ṇ), শ / ষ / স (ś / ṣ / s), র / ড় / ঢ় (r / ṛ / ṛh), etc. are prominent among even the native speakers. Spelling errors are further classified into 2 types.

1. **Non-Dictionary Words**: Spelling errors of this type result in words that are not in a dictionary. We have considered Vācaspati (Bhattacharyya et al., 2023) as the vocabulary of Bangla words since it covers literary works of almost 8 centuries and works from both India and Bangladesh. In the example shown in Table 1, কাজ (kāja) gets changed to কাব (kāva) which is not a word.
2. **Dictionary Words**: A spelling error of this type produces another word which is in the dictionary. However, in the context of the sentence, it is an error. For example, in Table 1, changing ড় (ṛ) of বাড়ি (vāṛi) to র (r) produces a perfect word বারি (vāri). The sentence, however, ceases to have any valid meaning. Mostly these errors are of *homonym* types, i.e., similar sounding words. Simple non-homonym typos may, however, also result in a dictionary word শাড়ি (śāṛi) that does not make sense in the sentence, as shown in the second example.

### D.2 Word Errors

A prominent class of grammatical errors in almost any language including Bangla is word errors. We have categorized word errors further into different sub-classes as explained next.

1. **Tense Error**: In Bangla, like most other languages, there are specific verb forms for the three tenses. Failing to use the correct form leads to errors, as illustrated in the example in Table 1. Tense errors are particularly prevalent when multiple verbs are used within a single sentence, resulting in mismatches among the verb tenses. In the second example in the table, while the first verb আসবে (āsavē) is in future tense, the second verb ফুটেছিল (phuṭēchila) is in past tense.
2. **Person Error**: Similar to tenses, there are different verb forms and pronouns for different persons in Bangla. It is, thus, an error to use the wrong person of a verb. The sentence in Table 1 shows an example where instead of the first person form করি (kari), the third person form করে (karē) is used with the pronoun আমি (āmi, I). These errors are common in Indian languages.
3. **Number Error**: In Bangla, the verb forms for both singular and plural numbers are the same. However, there are distinct forms for pronouns. The example in Table 1 shows such a wrong usage where the singular form আমি (āmi) is used instead of the plural form আমরা (āmarā). Number errors are more common in other Indian languages compared to Bangla.
4. **Gender Error**: In Bangla, the verb forms and pronouns for different genders are the same. However, there are distinct forms for adjectives as well as nouns. Moreover, the gender and number of an the adjective should match that of the noun it qualifies. Hence, in the example in Table 1, since the proper noun উত্তম (uttama) is masculine, the correct adjective used should be the masculine form অভিনেতা (abhinētā) and not the feminine form অভিনেত্রী (abhinētrī). While strictly speaking, masculine forms of adjectives should not be used for feminine nouns, it is a common practice to accept them. In such sentences, the masculine form takes the role of a gender-neutral form. Hence, the sentence সুচিত্রা একজন অসাধারণ অভিনেতা। (sucitrā ēkajana asādhāraṇa abhinētā., Suchitra is an outstanding actor.) where সুচিত্রা (sucitrā) is a feminine proper noun, but the adjective অভিনেতা (abhinētā) is masculine is not considered as incorrect. Many Indian languages, such as Hindi, have different forms of verbs for different genders and, thus, this kind of error is more common in those languages as compared to Bangla.
5. **Case Error**: Bangla and other Indian languages use a lot of inflected words. For different cases, different word forms are used that modify the original word. Case endings loosely correspond to prepositions in English. In the example in Table 1, the wrong case accusative is used instead of the correct case locative.
6. **Parts-of-Speech Error**: Sometimes, a word is used in a wrong parts-of-speech (POS). Since Indian languages, including Bangla, use a lot of nouns and their corresponding adjectives, these errors are common. Instead of a noun form, the adjective form is sometimes erroneously used, as shown in the example in Table 1.
7. **Missing Word Error**: These sentences are incomplete because of a missing word. Missing

a verb in Bangla will always generate this kind of error, as shown in the example in Table 1, while missing a random word may or may not be grammatically wrong. Missing a noun corresponding to its adjective will also generate an erroneous sentence. The second example in Table 1 shows such a sentence.

### D.3 Mixing of Language Variants: Gurucaṇḍālī Dōṣa

Bangla has a unique temporal language feature. All written works in Bangla till the 19th century were exclusively in সাধু ভাষা (sādhu bhāṣā, "refined language"). Authors started switching to চলিত ভাষা (calita bhāṣā, "colloquial language") during the 20th century and, currently, almost all the works are in this variant of the language. The two differ mostly in verb forms and pronouns and use exclusive sets of these. It is similar to the old English usage of "thou shalt" versus the modern "you shall", etc., but is more elaborate. A sentence should be written in either of the variants. Thus, it is an error to mix, for example, pronouns of one variant with verbs of another. The example in Table 1 shows two cases. The sentence নন্দবাবু ইহা লক্ষ্য করেছেন। (nandavāvu ihā lakṣya karēchēna.) mixes the sādhu bhāṣā pronoun form ইহা (ihā) with the calita bhāṣā verb form করিয়াছেন (kariẏāchēna). Either the verb form or the pronoun can be corrected, as shown in the examples. This mixing error is known as "গুরুচণ্ডালী দোষ" (Gurucaṇḍālī dōṣa) in Bangla.

### D.4 Punctuation Errors

Punctuation errors occur due to the usage of wrong punctuation marks, absence of punctuation marks where needed, or spurious usage of punctuation marks. Thus, while a simple imperative sentence ends with a । (full-stop mark), putting ? (interrogative mark) results in an error, as shown in Table 1.

### D.5 Semantic Errors

Semantic error is a special class of error where the sentence's semantic meaning becomes inconsistent or fictitious in the real world. For example, consider the sentence মানস আকাশ খেতে ভালোবাসে। (mānasa ākāśa khētē bhālōbāsē.) which literally means "Manas loves to eat the sky." Although this sentence is grammatically correct as far as the usage of words, spellings, etc. are concerned, it is still considered a wrong sentence due to its semantics. Note that this is for ordinary usage in a language, and such sentences may be correct in science fiction or other fantasy novels. Table 1 shows two correct sentences corresponding to the above wrong one. While in the first example, the verb is modified, in the second, the noun is modified to produce a semantically meaningful sentence.

### D.6 Multiple Errors

These sentences suffer from multiple errors of the same category or a combination of different categories of errors. For example, the sentence আমরা বান্ধুরা গতকাল কাশ্মীর যাবো। (āmarā bāndhurā gatakāla kāśmīra yābō., We <non-word> will go to Kashi tomorrow.) consists of spelling errors (non-dictionary), person errors and tense errors. The correct sentence can be আমরা বন্ধুরা আগামীকাল কাশী যাব। (āmarā bandhurā āgāmīkāla kāśī yāba।, We friends will go to Kashi tomorrow.)

## E Manual Generation

Table A2 provides the details of the essays given for the manual annotation survey. All 9 essays in this survey are commonly asked in 10th standard board exams. Each participant was asked to write an essay on a randomly picked topic. 36 participants undertook the study. Each annotator was paid on an hourly basis according to the standard rates prescribed by the university.

## F Methodology

Table A3 shows example sentences generated by back-translating English sentence to Bangla, whereas and Table A4 shows example sentences generated by round-trip translation with English as bridge language.

## G Generation of Different Types of Errors

- **Spelling Errors:** Spelling errors are those for which the original intention was to write the correct word, but some characters are wrongly written. Typically, the misspelled word should be within one or at most two edit distance from the original word. They can be, thus generated by substituting, inserting, or deleting one or two characters of a randomly chosen word in a sentence. These generated spelling errors may be of non-dictionary or dictionary types. We further collected 300 homonym word pairs from (Chakroborty, 2018). These homonyms are very common in Bangla. We replaced the original word in sentences with their corresponding homonyms to generate dictionary-based spelling errors.

- **Word Errors:** We have followed different procedures to generate different types of word errors in Bangla.

  1. **Tense Error:** We collected 24 most commonly used verbs and their forms across three tenses and three persons, resulting in 470 verb forms from (Chakroborty, 2018). This verb forms are replaced against the original word to generate erroneous sentences. These errors are difficult to generate since if the sentence contains only one verb and it is in its present form, then replacing it with past or future tense will not generate an error. For example in the sentence আমি বাড়ি যাব। (āmi vāṛi yāva.) (I will go home) if we replace "যাব" (yāva) with "গিয়েছিলাম" (giẏēchilāma) the resulting sentence আমি বাড়ি গিয়েছিলাম। (āmi vāṛi giẏēchilāma.) (I went to home) is not grammatically incorrect. So, to generate Tense errors, we need at least two verbs in the sentence, and we need to change only one, which we did randomly to generate Vaiyākaraṇa. For example, রাম যখন পড়তে বসবে, তখন লিখবে। rāma yakhana paṛatē vasavē, takhana likhavē. (Ram will write when he sits down to read.) if we change one among the verbs "বসবে" (vasavē) or "লিখবে" (likhavē) the resulting sentences রাম যখন পড়তে বসবে, তখন লিখেছিল। rāma yakhana paṛatē vasavē, takhana likhēchila. (Ram wrote when he sits down to read.) and রাম যখন পড়তে বসেছিল, তখন লিখবে। rāma yakhana paṛatē vasēchila, takhana likhavē. (Ram will write when he sat down to read) are grammatically wrong. If we change both the verbs, it will again result in a grammatically correct sentence রাম যখন পড়তে বসেছিল, তখন লিখেছিল। (rāma yakhana paṛatē vasēchila, takhana likhēchila.) (When Ram sat down to read, he wrote.) These issues compelled us to adopt the noise injection methodology to generate erroneous sentences, which would not have been possible if we had adopted other methodologies discussed in Sec: 2.There is one more type of tense error, which occurs with respect to time, like for sentence গতকাল আমি বাড়ি গিয়েছিলাম। (gatakāla āmi bāṛi giẏēchilāma.) (I went to home yesterday) if changed to গতকাল আমি বাড়ি যাব। gatakāla āmi bāṛi yāba. (I will go to home yesterday) or আগামীকাল আমি বাড়ি গিয়েছিলাম। āgāmīkāla āmi bāṛi giẏēchilāma. (I went to home tomorrow) will generate grammatically incorrect sentences. We have crafted 15 such sentences manually and added them to Vaiyākaraṇa.
  2. **Person Error:** To generate these types of errors, we replaced the original verb form with its corresponding verb form from the other two types of persons.
  3. **Number Error:** To generate this kind of error, we collected 23 pronouns with both of their singular-plural forms from (Chakroborty, 2018). We injected this error by deliberately replacing the original singular (respectively, plural) pronoun with its corresponding plural (respectively, singular) form. For pronoun detection, we used the POS tagger by Sarker (2021) since pronouns are typically a frozen list and taggers do well at detecting them.
  4. **Gender Error:** We handcrafted 100 sentences for this kind of error. In each sentence, we chose a random word and changed its case. We employed three native speakers to validate the error category, and based on majority voting, we added the sentences in Vaiyākaraṇa
  5. **Case Error:** We collected a list of 20 cases and inflections and randomly interchanged them in the sentences to generate the wrong sentences.
  6. **POS Error:** We collected 350 noun-adjective word pairs from (Chakroborty, 2018). We replaced a noun (respectively, adjective) with its corresponding adjective (respectively, noun) to generate errors.
  7. **Missing Word Error**: We ran the POS tagger (Sarker, 2021) and deleted verbs from the sentence to generate erroneous sentences. We applied the same technique to delete the noun corresponding to its adjective to generate errors. For other cases, we randomly deleted some words from the sentences. We asked three native speakers to validate whether the generated sentence was an error, and based on majority voting, we marked the sentences. If it is an error, we add the sentence to Vaiyākaraṇa. Else, we discard it.

- **Semantic Error:** We handcrafted 100 sentences for this kind of error. We employed three native speakers to validate the error category of the sentences, and based on majority voting, we added the sentences in Vaiyākaraṇa.
- **Gurucaṇḍālī Dōṣa:** We collected 140+ verbs and pronouns with their corresponding sādhu and calita forms from (Chakroborty, 2018). We then replaced the original word with its counterpart to generate this kind of error. To generate these sentences, we make sure that at least one verb or pronoun retains its original form so that the resulting sentence is an error that mixes the two variants. For example আমি খেতে খেতে হাঁটছি। (āmi khētē khētē hām̐ṭachi.) we randomly changed "খেতে" (khētē) to "খাইতে" (khāitē) generating a wrong sentence আমি খাইতে খেতে হাঁটছি। (āmi khāitē khētē hām̐ṭachi.). Changing all three "খেতে" (khētē), "খেতে" (khētē) and "হাঁটছি" (hām̐ṭachi) will lead to a grammatically correct sentence আমি খাইতে খাইতে হাঁটিতেছিলাম। (āmi khāitē khāitē hām̐ṭitēchilāma.).

Following these steps, we generated 2,26,869 grammatically incorrect sentences, as outlined in Table 3. Following the procedure outlined above, we can generate any number of grammatically incorrect sentences for Bangla.

## H Data Cleaning

- *Cleaning of Unicode characters*: Unicode characters "0020" (space), "00$a$0" (no-break space), "200$c$" (zero width non-joiner), "1680" (ogham space mark), "180$e$" (mongolian vowel separator), "202$f$" (narrow no-break space), "205$f$" (medium mathematical space), "3000" (ideographic space), "2000" (en quad), "200$a$" (hair space) are separated from the texts.
- *Cleaning of different punctuation marks*: In Bangla, usage of punctuation marks has also evolved alongside words. In particular, we have treated the following as punctuation marks: "...", "।...", "।।", "!–", "–".

## I Statistical Comparison of Real and Synthetic Data

In this section, we assess whether the distribution of error categories in the manual analysis is similar to that in Vaiyākaraṇa. We conducted this validation in three different ways. First, we performed a *binary* task to compare the distribution of correct and incorrect sentences in both the manual annotation and Vaiyākaraṇa. The second task involved a *multi-class* validation across five broad error categories (excluding sub-classes) and their respective distributions. Lastly, we examined the distribution of finer classes (totalling 12) between the manual analysis and Vaiyākaraṇa. For these validation tasks, we employed the Jensen-Shannon divergence and will discuss the significance of our findings.

Table A5 presents the JSD values for all three tasks, which are all less than 0.05. This finding suggests that the distribution of erroneous sentences and error categories in the manual data and Vaiyākaraṇa are comparable. Therefore, we can use Vaiyākaraṇa as a benchmark dataset for Grammatical Error Correction (GEC) in Bangla.

## J Multiple Error Categories

We have observed from the manual survey that there are single errors in a sentence 77% of the time. The remaining 23% of the sentences contain multiple errors. Amongst them, the most common is multiple spelling errors in a sentence. We have categorised multiple errors of the same type in that category of error only, not inducted them in the multiple error category. Vaiyākaraṇa has 3,860 such sentences. Very few sentences have more than one type of error. Here, we show the distribution of sentences having different types of errors in a sentence. We have only considered that category of errors prevalent in the human survey in Sec. 4. Table A6 show the distribution of multiple errors in Vaiyākaraṇa.

## K Results of Transformer-based Models

Table A7 shows the performance of all neural models and Random Forest classifier on the 5,67,422 sentences in Vaiyākaraṇa.

We also evaluated decoder-based models to detect whether a sentence is grammatically correct. Table A8 presents the performance of decoder-based models for prompt with prompt "বাক্যটি সঠিক অথবা ভুল কিনা তা নির্ধারণ কর।" (bākyaṭi saṭhika athabā bhula kinā tā nirdhāraṇa kara., is this sentence grammatically correct?). Since the performance of binary classification is itself poor we have not continued with the multi-class classifications.

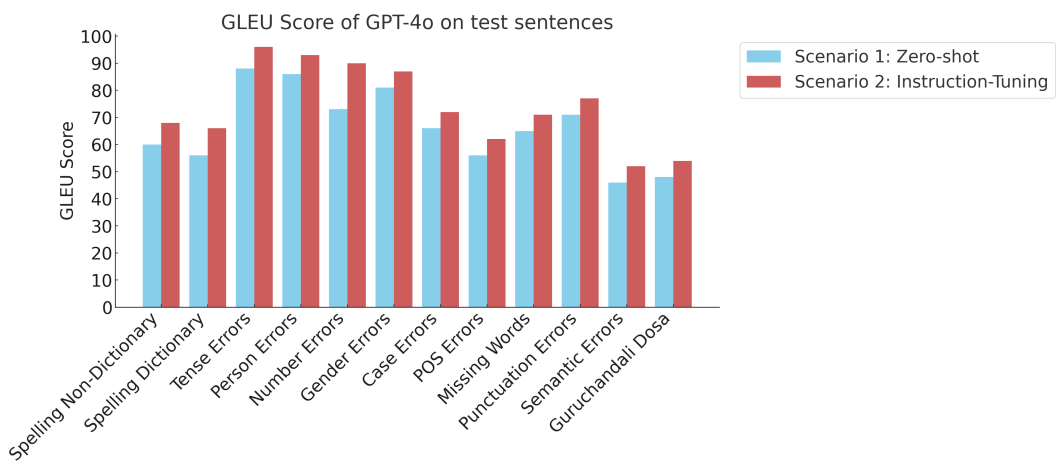

Figure 3: Figure showing performance of GPT-4o on different error categories in Bangla on 20,100 generated erroneous sentences.

## L Grammatical Error Correction

We created a test set of 52,100 sentences to evaluate the efficacy of Vaiyākaraṇa. The distribution of error categories on the test set is shown in Table A9.

Fig. 3 shows the GLEU score for each of the 12 error categories for GPT-4o on the 52,100 sentences used for GEC evaluation of transformer models.

## M Model Hyper Parameters

We fine-tuned the transformer-based models with Adam optimizer (Kingma and Ba, 2015) and learning rate of 2e-5 for 20 epochs. Each transformer-based model's batch size is 16, with a maximum length of 512.

## N Hyperparameters for Instruction-tuning

All models were instruction-tuned using the Low-Rank Adaptation (LoRA) method (Hu et al., 2021), a parameter-efficient fine-tuning approach for pre-trained models (Xu et al., 2023), via the Hugging Face *peft* module. The LoRA hyperparameters were set as follows:

- Rank (r): 16
- LoRA alpha ($\alpha$): 32
- LoRA dropout: 0.05
- Bias: none

All other hyperparameters were set at their default values. For all the models, the following hyperparameter values have been used for generation:

- temperature=0.7
- top_k=50
- num_beams=10
- max_length=1800

The default values have been used for all other hyperparameters.

## O Anecdotal Examples

Table A10 shows a few anecdotal examples of Human and GPT-4o (best-performing model) generated outputs on test sentences from Vaiyākaraṇa dataset. In the first example, even though the generated output is grammatically correct, the edit distance between the generated sentence and ground truth is significant, resulting in a decreased GLEU score. The generated output and ground truth are the same in the second case. Hence, the GLEU score is 1.0.

| Essay topic | # Essays |
|---|---|
| বিজ্ঞান আশীর্বাদ না অভিশাপ<br>vijñāna āśīrvāda nā abhiśāpa<br>Is science boon or bane? | 15 |
| একটি বৃষ্টির দিন<br>ēkaṭi vr̥ṣṭira dina<br>A rainy day | 12 |
| একটি নদীর আত্মকথা<br>ēkaṭi nadīra ātmakathā<br>Autobiography of a river | 6 |
| একটি স্মরণীয় দিন<br>ēkaṭi smaraṇīẏa dina<br>A memorable day | 15 |
| খেলা শুধু খেলা নয়<br>khēlā śudhu khēlā naẏa<br>Sports is not just sports | 12 |
| হঠাৎ আলাদিনের আশ্চর্য প্রদীপ কুড়িয়ে পেলে কী করবে<br>haṭhāt ālādinēra āścarya pradīpa kuṛiẏē pēlē kī karavē<br>What will you do if you suddenly get Al-addin's lamp? | 8 |
| সামাজিক মাধ্যম আশীর্বাদ না অভিশাপ<br>sāmājika mādhyama āśīrvāda nā abhiśāpa<br>Is social media boon or bane? | 20 |
| ভীন গ্রহের প্রাণী ও তোমার কথোপকথন<br>bhīna grahēra prāṇī ō tōmāra kathōpakathana<br>Dialogue between an extraterrestrial being and you | 10 |
| পনেরো বছর আগের তুমি আর আজকের তুমির মধ্যে কথোপকথন<br>panērō vachara āgēra tumi āra ājakēra tumira madhyē kathōpakathana<br>Dialogue between 15-years older you and present you | 15 |
| একটি বটগাছের আত্মকথা<br>ēkaṭi vaṭagāchēra ātmakathā<br>Autobiography of a Banyan Tree | 10 |
| **Total** | **123** |

Table A2: Grammar essays for manual survey

| English | Bangla | Grammar |
|---|---|---|
| *He helped me succeed.* | তিনি আমাকে সফল করতে সাহায্য করেছেন। (tini āmākē saphala karatē sāhāyya karēchēna.) | Semantic |
| *I go to school.* | আমি স্কুলে যাই। (āmi skulē yāi.) | Correct |

Table A3: Example of sentences generated by back translation from English to Bangla.

| Bangla | English | Bangla | Grammar |
|---|---|---|---|
| সে আমার বন্ধু। (sē āmāra bandhu.) | He is my friend | সে আমার বন্ধুটা। (sē āmāra bandhuṭā.) | Case |
| নন্দবাবু এটা লক্ষ্য করলেন। (nandabābu ēṭā lakṣya karalēna.) | Nandababu noticed this. | নন্দ বাবু ব্যাপারটা লক্ষ্য করলেন। (nanda bābu byāpāraṭā lakṣya karalēna.) | Correct |

Table A4: Example of sentences generated by round-trip translation using English as bridge language.

| Type | JSD-Score |
|---|---|
| Binary | 0.048 |
| Broader | 0.037 |
| Finer | 0.046 |

Table A5: Jensen-Shannon Divergence score of distribution of manual analysis and Vaiyākaraṇa.

| Error Class | #Occurrences |
|---|---|
| Non_Dictionary-Dictionary | 1,860 |
| Case-Non_Dictionary-Dictionary | 500 |
| Gurucaṇḍālī Dōṣa-Non_Dictionary-Dictionary | 250 |
| Non_Dictionary-Dictionary-Punctuation | 250 |
| Punctuation-Punctuation | 250 |
| Case-Non_Dictionary-Dictionary-Gurucaṇḍālī Dōṣa | 250 |
| Non_Dictionary-Dictionary-Gurucaṇḍālī Dōṣa-Punctuation | 250 |
| Non_Dictionary-Dictionary-Gurucaṇḍālī Dōṣa-Punctuation-Case-Missing Word | 250 |
| Total | 3,860 |

Table A6: Distribution of Multiple Errors in Vaiyākaraṇa

| Model | Parameters | Binary | Broad | Finer |
|---|---|---|---|---|
| Google-ByT5 | 300M | 81.25 ±0.65 | 79.30±0.87 | 76.65±0.80 |
| BanglaT5 | 270M | 88.90±0.10 | 84.50±0.68 | 82.48±0.14 |
| Panini | 70.46M | 89.25±0.25 | 84.75±0.15 | 82.88±0.08 |

Table A7: Macro-F1 of transformer models on Vaiyākaraṇa.

| Model | Without Instruction Tuning | With Instruction Tuning |
|---|---|---|
| GPT-4.0 | 40.30 | 56.30 |
| GPT-2-XL | 38.10 | 49.65 |
| OPT | 36.50 | 48.30 |
| GPT-Neo | 36.38 | 50.90 |
| BLOOM | 35.00 | 48.00 |

Table A8: Macro-F1 of different LLMs for prompts with and without instruction tuning for 600 sentences.

| Error Class | #Occurrences |
|---|---|
| Non-Dictionary | 8,000 |
| Dictionary | 2,000 |
| Spelling Errors | 10,000 |
| Tense Errors | 1,000 |
| Person Errors | 1,000 |
| Number Errors | 40 |
| Gender Errors | 30 |
| Case Errors | 2,000 |
| POS Errors | 750 |
| Missing Words | 750 |
| Word Errors | 5,570 |
| Punctuation Errors | 2,000 |
| Semantic Errors | 30 |
| Gurucaṇḍālī Dōṣa | 2,000 |
| Multiple Errors Different Categories | 500 |
| *InCorrect* | 20,100 |
| *Correct* | 32,000 |
| **Total** | 52,100 |

Table A9: Distribution of different error categories in the test set of 52,100 sentences.

| Test Sentence | Correct | Human | GPT-4o (Zero-Shot) | GPT-4o (Instruction Tuned) |
|---|---|---|---|---|
| কানে গাইতেন বাজতেই লাগল। (kānē gāitēna bājatēi lāgala., incorrect sentence) | কানে গানটা বাজতেই লাগল। (kānē gānaṭā bājatēi lāgala., The song kept ringing in my ears.) | কানে গানটা বাজতেই লাগল। (kānē gānaṭā bājatēi lāgala., The song kept ringing in my ears.) | কানে গাইতেন বাজতেই লাগল। (kānē gāitēna bājatēi lāgala., incorrect sentence) | কানে ইয়ারফোন পরতেই গান বাজতে লাগল। (kānē iẏāraphōna paratēi gāna bājatē lāgala., As soon as I put on the earphones, the music started playing) |
| শিস দিতে গিয়ে দেখলেন, শব্দ হয়েছিল না তো! (śisa ditē giẏē dēkhalēna, śabda haẏēchila nā tō!, incorrect sentence) | শিস দিতে গিয়ে দেখলেন, শব্দ হল না তো! (śisa ditē giẏē dēkhalēna, śabda hala nā tō!, When he whistled, he saw that there was no sound!) | শিস দিতে গিয়ে দেখলেন, শব্দ হল না তো! (śisa ditē giẏē dēkhalēna, śabda hala nā tō!, When he whistled, he saw that there was no sound!) | শিস দিতে গিয়ে দেখলেন, শব্দ হয়নি তো! (śisa ditē giẏē dēkhalēna, śabda haẏani tō!, wrong use of verb form) | শিস দিতে গিয়ে দেখলেন, শব্দ হল না তো! (śisa ditē giẏē dēkhalēna, śabda hala nā tō!, When he whistled, he saw that there was no sound!) |

Table A10: Anecdotal examples comparing human versus GPT-4o at both zero-shot and instruction-tuned